\documentclass[10pt,twocolumn,letterpaper]{article}
\usepackage{multirow}
\usepackage[pagenumbers]{cvpr} 
\usepackage[accsupp]{axessibility}

\usepackage[dvipsnames]{xcolor}

\definecolor{cvprblue}{rgb}{0.21,0.49,0.74}
\usepackage[pagebackref,breaklinks,colorlinks,citecolor=cvprblue]{hyperref}
\usepackage{algpseudocode}
\usepackage{algorithm}
\def\paperID{*****} 
\def\confName{CVPR}
\def\confYear{2024}

\title{Plug-and-Play Acceleration of Occupancy Grid-based NeRF Rendering using VDB Grid and Hierarchical Ray Traversal}

\author{Yoshio Kato, Shuhei Tarashima \\
NTT Communications, Japan\\
{\tt\small \{yoshio.kato, shuhei.tarashima\}@ntt.com}
}

\begin{document}
\maketitle
\begin{abstract}
Transmittance estimators such as Occupancy Grid (OG)\cite{ingp} can accelerate the training and rendering of Neural Radiance Field (NeRF) by predicting important samples that contributes much to the generated image. 
However, OG manages occupied regions in the form of the dense binary grid, in which there are many blocks with the same values that cause redundant examination of voxels' emptiness in ray-tracing. In our work, we introduce two techniques to improve the efficiency of ray-tracing in trained OG without fine-tuning. First, we replace the dense grids with VDB\cite{vdb} grids to reduce the spatial redundancy. Second, we use hierarchical digital differential analyzer (HDDA)\cite{hdda} to efficiently trace voxels in the VDB grids. Our experiments on NeRF-Synthetic\cite{nerf} and Mip-NeRF 360\cite{mipnerf360} datasets show that our proposed method successfully accelerates rendering NeRF-Synthetic dataset\cite{nerf} by 12\% in average and Mip-NeRF 360 dataset\cite{mipnerf360} by 4\% in average, compared to a fast implementation of OG, NerfAcc\cite{nerfacc}, without losing the quality of rendered images.
\end{abstract}
    
\section{Introduction}
\label{sec:intro}

Neural Radiance Fields (NeRFs) have shown impressive results in novel view synthesis tasks by using neural networks and ray tracing to model volume density and view-dependent appearance\cite{nerf}. While NeRF-based methods represent detailed geometry and photorealistic appearance, they are suffered from high computational costs in rendering. For instance, the original NeRF method requires days to train a single scene and 10 seconds to render a frame\cite{nerf}.

Usually a NeRF model can be divided into two main modules: a ray-sampler and a radiance field. When we calculate a color of the given pixel, at first the ray-sampler decides the sampling points on the camera ray that associates to the pixel, and then sampled points are fed into the radiance field to predict the density and color of each point. Finally, these values are accumulated to the pixel color.
Recently, many methods for accelerating training or rendering have been proposed by modifying radiance fields\cite{ingp, dvgo, plenoxels, tensorf, kplanes, hexplane} or improving ray-samplers\cite{ingp, mipnerf360}. In addition, it is known that they also improve convergence speed in training.
In this work, we focus on ray-samplers to give benefits to the trained NeRF models without large computational cost such as fine-tuning\cite{plenoctrees}.
Since improving ray-samplers and radiance fields orthogonally contribute to rendering acceleration, our work on ray-samplers can be combined with any Occupancy Grid (OG)\cite{ingp}-based model.

The main aim of ray-samplers is to sample important points on a ray. In ray-tracing, we can reduce the number of samples without losing the quality by selecting points with higher densities. For example, NerfAcc\cite{nerfacc} demonstrates that OG can serve as a good sampling method of both faster convergence and rendering for various NeRF methods.

Dispite its performance and usability, OG manages occupied regions in the form of the dense binary grid and has spatial redundancy. It makes rendering slower due to a number of unnecessary operations to search occupied voxels on a ray. To resolve this problem, we propose a method that replace the OG-based ray-sampler without re-training or fine-tuning. Our method is composed of two techniques. First, to reduce spatial redundancy in OG, we use VDB\cite{vdb}, a data structure for sparse volumes. This data structure merges a block of a number of empty voxels. Second, to synergistically accelerate sampling with VDB, we use an efficient ray-tracing algorithm called Hierarchical Digital Differential Analyzer (HDDA)\cite{hdda}, which skips a merged empty voxel at one operation. These two techniques reduce the number of operations to traverse voxels on a ray, and accelerate the ray-sampler.

As far as we know, this work is the first paper that utilizes a hierarchical data structure and a hierarchical ray tracing algorithm simultaneously in the context of NeRF rendering acceleration.
To evaluate our method, we trained Instant-NGP\cite{ingp} with a normal OG-based ray-sampler and replaced it with our VDB-based algorithm before starting to render images. We used NeRF-Synthetic dataset\cite{nerf} and Mip-NeRF 360 dataset\cite{mipnerf360}, and showed that in most cases our method improved rendering FPS while preserving PSNR compared to the NerfAcc\cite{nerfacc} implementation. In addition, we showed using VDB and HDDA together makes much larger improvements than using either technique in some cases. Our code is available at \url{https://github.com/Yosshi999/faster-occgrid}.

\section{Related Works}
\label{sec:relwork}

\subsection{Efficient Ray-Sampler}
By sampling the points which contributes much to the final rendered image, \textit{i.e.}, the points with high density, we can reduce the number of samples without losing the quality of synthesized images. Several approaches to identify high-density points have been proposed. Occupancy Grid (OG)\cite{ingp} caches the predicted densities in NeRF output during training, and the ray tracer samples from the voxels whose density is higher than a threshold. Mip-NeRF 360\cite{mipnerf360} proposed Proposal Network (PN), a neural network to estimate the density of queried points. However, since PN requires a number of network forward inferences to suggest sampling points, it unavoidably demands additional computational costs. Therefore, PN contributes to faster convergence of training but is not suited to improve rendering efficiency. In this work, we adopt OG as a baseline because we focus on the faster rendering.

\subsection{Sparse Data Structures}
Some NeRF methods create feature grids to represent a radiance field, in which there are usually large vacuum regions. Thus, it is important to reduce the number of accesses to such vaccum regions. This is also true for OG. To reduce such spatial redundancy in the grid, we can adopt sparse data structures that are widely investigated in the domain of Computer Graphics (CG). One approach is Octree\cite{octree}, which recursively subdivides the grid and represents its containment relationship as a tree. It can reduce its spatial redundancy by pruning the nodes which associate to the empty spaces. This enables skipping large empty regions and accelerating ray-tracing. Instant-NGP\cite{ingp} uses Octree-like structure when it represents OG in training larger scenes, but no pruning is performed.
Another approach is VDB\cite{vdb}, a tree-like data structure to store sparse voxels in a hierarchical manner. Unlike Octree, VDB fixes the depth of its tree and saves the computational cost to 
traverse the tree in accessing voxels. In addition, VDB provides a caching mechanism to accelerate accessing neighboring voxels. Notice that PlenVDB\cite{plenvdb} already adopts VDB for representing the radiance field to achieve fast voxel accessing and efficient memory usage. However, it doesn't utilize VDB's hierarchical structure for ray-tracing. This fact motivated us to fully exploit VDB for OG, which is connectable to various NeRF-based methods. In addition, we focus on more efficient ray-tracing algorithms that could be run on VDB's hierarchical structure, which will be described in the next section.

\subsection{Enumerating Line-Grid Intersections}
\label{sec:rel3}
In CG research community, efficient ray marching methods are widely investigated. They are also useful for faster NeRF rendering. In this section, we introduce some ray marching algorithm that suit with grid based NeRF algorithms.

\begin{figure}
\centering
\includegraphics[width=0.9\linewidth]{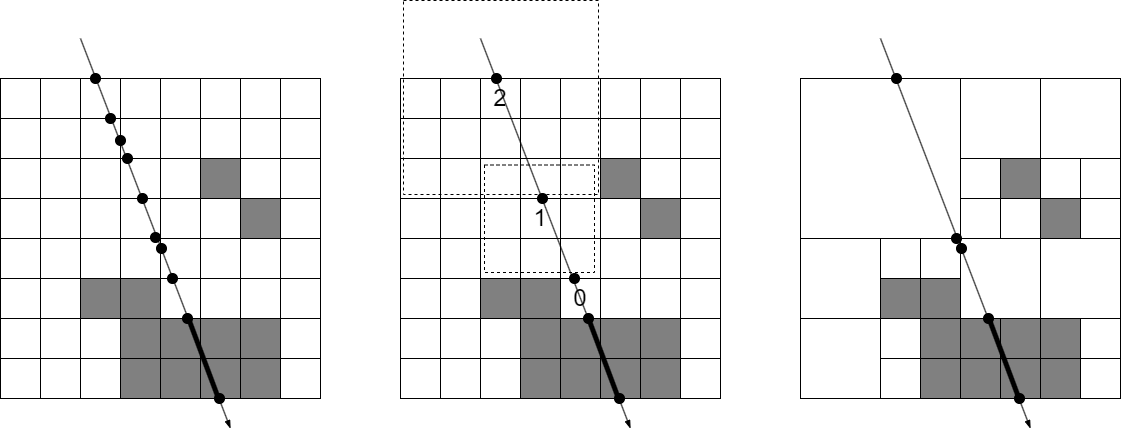}
\caption{Examples of ray-marching algorithms. DDA (left), Chessboard Distance (CD)\cite{cd} (center), and HDDA on quadtree (right). Black dots depict lookup operations. Black lines are where the colors are sampled. White voxel means empty space. In DDA, we need lookup the voxel's occupancy one by one. CD saves the $L_\infty$ distance to the nearest occupied voxel in advance. HDDA can skip by power of two according to the information in the quadtree.}
\label{fig:grid}
\end{figure}

A widely adopted algorithm is DDA\cite{dda}. It efficiently enumerates the voxels that intersect a given line. In NeRF rendering, we can sample points from occupied voxels on a ray while skipping empty voxels. However, NeRF scenes usually have many empty regions and skipping only one voxel for each iteration is inefficient.
There have been some approaches to efficiently skip multiple voxels in the literature.
One way is to retrieve the distance for which we can skip safely from the current point\cite{cd, largejump}. However, this requires large additional memory to save integers in the all voxels. Another approach is a hierarchical technique. HDDA\cite{hdda} utilizes VDB's hierarchical structure and skips a large empty subspace. \cref{fig:grid} visualizes three approaches: DDA\cite{dda} as a baseline, Chessboard Distance\cite{cd} as a skipping method with distance retrieval, and HDDA\cite{hdda} as a hierarchical approach. In this work, we use HDDA to efficiently search significant voxels in our VDB-structured OG. As far as we know, modifing the ray-tracing algorithm to suit hierarchical data structure is the first attempt in the domain of accelerating NeRF's ray-sampler.

\begin{table*}[bt]
  \centering
\begin{tabular}{c|cccccccc|c}
\toprule
PSNR & chair & drums & ficus & hotdog & lego & materials & mic & ship & ave. \\
\midrule
NerfAcc & 35.74 & 25.45 & 33.96 & 37.33 & 35.79 & 29.58 & 36.71 & 30.57 & 33.14 \\
VDB + DDA branch & 35.74 & 25.45 & 33.96 & 37.33 & 35.79 & 29.58 & 36.71 & 30.57 & 33.14 \\
VDB + DDA skip & 35.74 & 25.45 & 33.96 & 37.33 & 35.79 & 29.58 & 36.71 & 30.57 & 33.14\\
VDB + HDDA branch & 35.64 & 25.43 & 33.96 & 37.33 & 35.68 & 29.57 & 36.64 & 30.55 & 33.10 \\
VDB + HDDA skip & 35.64 & 25.43 & 33.96 & 37.33 & 35.68 & 29.57 & 36.64 & 30.55 & 33.10 \\
\end{tabular}

\begin{tabular}{c|cccccccc|cc}
\toprule
FPS & chair & drums & ficus & hotdog & lego & materials & mic & ship & ave.  \\
\midrule
NerfAcc & 4.22 & 4.19 & 4.27 & 4.24 & 4.29 & 5.06 & 4.48 & 3.28 & 4.25 \\
VDB + DDA branch & 4.67 & \bf 4.58 & \bf 4.63 & 4.89 & 4.77 & 5.63 & 4.86 & 3.78 & 4.73 \\
VDB + DDA skip & 4.03 & 3.83 & 3.94 & 4.32 & 4.08 & 4.79 & 4.29 & 3.46 & 4.09 \\
VDB + HDDA branch & 4.60 & 4.32 & 4.41 & 4.86 & 4.76 & 5.68 & 4.85 & 3.71 & 4.64 \\
VDB + HDDA skip & \bf 4.73 & 4.48 & 4.54 & \bf 4.94 & \bf 4.82 & \bf 5.85 & \bf 5.03 & \bf 3.80 & \bf 4.77 \\
\hline
\bottomrule
\end{tabular}
  \caption{Rendering results on NeRF-Synthetic dataset.}
  \label{tab:nerfcomp}
\end{table*}

\begin{table*}[bt]
  \centering
\begin{tabular}{c|ccccccccc|c}
\toprule
\multirow{2}{*}{PSNR} & \multicolumn{5}{c|}{outdoor} & \multicolumn{4}{c|}{indoor} & \\
& bicycle & flowers & garden & stump & treehill & room & counter & kitchen & bonsai  & ave. \\
\midrule
NerfAcc & 22.33 & 19.98 & 24.56 & 23.17 & 22.06 & 30.52 & 26.90 & 27.92 & 29.99 & 25.27 \\
VDB + DDA branch & 22.33 & 19.98 & 24.56 & 23.17 & 22.06 & 30.52 & 26.90 & 27.92 & 29.99 & 25.27\\
VDB + DDA skip & 22.34 & 19.98 & 24.56 & 23.17 & 22.06 & 30.53 & 26.90 & 27.93 & 30.00 & 25.28 \\
VDB + HDDA branch & 22.32 & 19.98 & 24.56 & 23.17 & 22.06 & 30.5 & 26.88 & 27.91 & 29.98 & 25.26 \\
VDB + HDDA skip & 22.32 & 19.98 & 24.56 & 23.17 & 22.06 & 30.51 & 26.88 & 27.92 & 29.99 & 25.27 \\
\end{tabular}
\begin{tabular}{c|ccccccccc|c}
\toprule
\multirow{2}{*}{FPS} & \multicolumn{5}{c|}{outdoor} & \multicolumn{4}{c|}{indoor} & \\
& bicycle & flowers & garden & stump & treehill & room & counter & kitchen & bonsai  & ave.\\
\midrule
NerfAcc & 1.21 & 1.64 & 1.63 & 1.17 & 1.30 & 4.37 & 4.03 & 4.61 & 4.20 & 2.69  \\
VDB + DDA branch & \bf 1.34 & \bf 1.80 & \bf 1.83 & \bf 1.29 & \bf 1.41 & \bf 4.64 & \bf 4.15 & \bf 4.78 & 4.36 & \bf 2.84 \\
VDB + DDA skip & 1.25 & 1.62 & 1.69 & 1.21 & 1.27 & 4.27 & 3.63 & 4.16 & 3.80 & 2.55 \\
VDB + HDDA branch & 1.31 & 1.78 & 1.81 & 1.27 & 1.37 & 4.53 & 4.11 & 4.74 & 4.39 & 2.81 \\
VDB + HDDA skip &1.31 & 1.80 & 1.82 & 1.28 & 1.39 & 4.59 & 3.96 & 4.66 & \bf 4.46 &2.81  \\
\hline
\bottomrule
\end{tabular}
  \caption{Rendering results on Mip-NeRF 360 dataset.}
  \label{tab:mipcomp}
\end{table*}

\section{Method}
\label{sec:method}
We convert Occupancy Grid (OG)\cite{ingp} of a trained NeRF model into VDB-based structure using OpenVDB\cite{openvdb}, and transfer it to GPU using NanoVDB\cite{nanovdb}. In this work, we use Instant-NGP\cite{ingp} as a base model. If there are some redundancies in the VDB grid, we prune them and manage the large occupied or unoccupied region as a tile, where the tile contains multiple voxels sharing same values. When the ray-sampler collects points on a ray, we use HDDA\cite{hdda} to efficiently traverse grids on the ray. Every time we find an occupied voxel or a fully occupied tile, we sample points along the ray at specified intervals; a constant stepsize for NeRF-Synthetic dataset\cite{nerf} and a linearly increasing stepsize for Mip-NeRF 360 dataset\cite{mipnerf360}. Usually ray-sampling is processed on GPU and each thread is assigned to each ray. We note that there are mainly two implementation variants for the ray-sampling kernel. One algorithm branches in traversing a grid according to its occupancy, and the other skips all empty cells and then marches in a grid. The former, we named it \textit{DDA branch}, is adopted by NerfAcc\cite{nerfacc}, and the latter, we named it \textit{DDA skip}, is adoped by Instant-NGP. We show the algorithms in the supplementary material \cref{sec:vari}. In this work, we compare the rendering speed by changing the ray-sampling kernel.

For larger scenes such as Mip-NeRF 360, we prepare multi-resolution VDB grids by following the setting of Instant-NGP\cite{ingp}. When the ray-sampler accesses the voxels and multiple voxels are intersected at the same time, it prioritises finer voxels.


\section{Experimental Results}

\subsection{Experimental Setup}
We used NeRF-Synthetic dataset\cite{nerf} for bounded scenes and Mip-NeRF 360 dataset\cite{mipnerf360} for unbounded scenes. NeRF-Synthetic contains eight synthetic scenes and each scene has 100 images for training and 200 images for testing. We prepared a single Occupancy Grid (OG) with resolution $128^3$ for NeRF-Synthetic. Mip-NeRF 360 contains four indoor and five outdoor photorealistic scenes and each scene has between 100 and 330. We followed the setting of Instant-NGP\cite{ingp} and prepared four multi-resolution grids and each grid has resolution $128^3$ and voxel size $2^b$, where $b \in \lbrack 0, 3 \rbrack$ for Mip-NeRF 360.
On both datasets we trained Instant-NGP with 20k iterations and the batchsize of 1024. We updated OG every 16 iterations in training. We used NerfAcc\cite{nerfacc} as a baseline and forked it to implement our VDB-based ray-sampler and experiments. As far as we know, NerfAcc is the fastest implementation of OG\cite{nerfacc}.

After training NeRF models, we converted the dense grids of OG into VDB\cite{vdb}. It took only 60 msec to covert the dense grids in the worst case. Next, we rendered the test images to measure FPS, PSNR, and the amount of memory allocated by VDB grids. We selected the ray-tracing algorithm from DDA and HDDA described in \cref{sec:rel3}, and selected the kernel implementation from \textit{DDA branch} and \textit{DDA skip} described in \cref{sec:method}.
All experiments are conducted on a Ubuntu server with 1x NVIDIA A100 GPU.

\subsection{Results}
\cref{tab:nerfcomp} shows that our method improves the rendering FPS in the all cases in NeRF-Synthetic dataset. In most cases, the ray-sampler using VDB and HDDA with the \textit{DDA skip} kernel records the fastest FPS, and accelerate rendering by 12\% in average. In the best case, \textit{hotdog} scene gets faster by 16\% in rendering. Note that no scene got significantly worse PSNR than the baseline.

\cref{tab:mipcomp} shows that in Mip-NeRF 360 dataset our method improves the rendering FPS by 4-5\% in average. Similarly to the experiment on NeRF-Synthetic dataset, all scenes don't get worse significantly in PSNR. We note that introducing HDDA did not work well for Mip-NeRF 360 dataset and in most cases the ray-sampler with DDA and the kernel \textit{DDA branch} was the best. In addition, the extent of speedup on Mip-NeRF 360 dataset is smaller than that of NeRF-Synthetic dataset. 

\cref{tab:nerfmem} shows that the memory usage of VDB grids is about twice larger than the baseline, which uses a bitarray representing dense boolean grids. In particular, VDB grids converted from Occupancy Grid trained on Mip-NeRF 360 dataset are tend to be much larger because of the VDB's overhead.
However, we believe it is acceptable because additional memory usage is at most 1.5MiB, since the size of a family of compact NeRF-based models is around 5MiB\cite{tensorf}.

\begin{table}[tb]
\begin{minipage}[t]{.45\linewidth}
  \centering
\begin{tabular}{c|c}
& Size (KiB)\\
\toprule
chair & 336.5\\
drums & 410.4 \\
ficus & 388.6 \\
hotdog & 337.5 \\
lego & 339.9 \\
materials & 327.4 \\
mic & 316.1 \\
ship & 363.1 \\
\midrule
ave. & 352.4 \\
\bottomrule
\end{tabular}
\end{minipage}
\begin{minipage}[t]{.45\linewidth}
 \centering
\begin{tabular}{c|c}
& Size (KiB)\\
\toprule
bicycle & 2259.0\\
flowers & 2145.4\\
garden & 2089.7\\
stump & 2157.9\\
treehill & 2279.8\\
room & 1999.1\\
counter & 2116.8\\
kitchen & 2155.8\\
bonsai  & 2105.8\\
\midrule
ave. & 2145.5 \\
\bottomrule
\end{tabular}
\end{minipage}
  \caption{The size of VDB grids converted from Occupancy Grid (OG) trained on NeRF-Synthetic dataset (left) and Mip-NeRF 360 dataset (right). Before conversion, OG trained on NeRF-Synthetic is 256KiB and OG trained on Mip-NeRF 360 is 1024KiB.}
  \label{tab:nerfmem}
\end{table}

\section{Discussion}
In this study we proposed to use a VDB-based Occupancy Grid (OG) and a sampling algorithm using HDDA to accelerate NeRF rendering. Our experiments showed that it improves rendering speed, without losing the quality of rendered images. We emphasize that our method can be adopted for the trained NeRF models whose ray-sampler is based on OG. However, for larger scenes such as Mip-NeRF 360\cite{mipnerf360}, our improvement tends to be marginal, possibly because of an overhead in processing multiple VDB grids. It is necessary to compare the amount of additional memory consumption and FPS improvement when our method is adopted in production. In addition, when we replaced the ray-sampler with ours, it was not able to generate the entirely same images as the original ray-sampler generates. One potential reason lies in floating errors that could negatively affects the rendering results.

We also find that \textit{DDA skip}, skipping leading empty cells in advance, did not work well with DDA, as shown in "VDB + DDA skip" rows of \cref{tab:nerfcomp} and \cref{tab:mipcomp}. However, by using VDB together with HDDA, the rendering speed got faster than the baseline.
While we believe this is because HDDA could reduce the number of operations for skipping voxels and saved the computational time in all threads, deeper investigations of the CUDA performances should be performed in the future work.

It is notable that using other sparse data structure such as Octree is also worth investigating. While VDB restricts the depth of tree and requires larger sub-grids for each level, the dimension of sub-grid is always $2^3$ in Octree and its fineness may help the effectiveness of representing sparse scenes. In addition, Octree-based OG may be more memory-efficient because the structure of Octree is simpler than VDB.

A possible extension of our work is to accelerate training. One of the difficulties of adopting efficient data structure in training is to minimize the overhead of optimizing spatial redundancy.
Since pruning requires computation time to some extent, we can't frequently prune the variables that is intermittently updated during training.
Our proposed method is not suitable for frequent updates because the VDB grid located on GPU is read-only and we have to re-transfer the grid pruned on CPU to GPU. An efficient backend with low-cost pruning is worth investigating.
\newpage

{
    \small
    \bibliographystyle{ieeenat_fullname}
    \bibliography{main}
}

\clearpage
\setcounter{page}{1}
\maketitlesupplementary

\section{Variants of Ray-Sampling Kernel}
\label{sec:vari}

\begin{algorithm}[tb]
    \caption{Algorithm of DDA branch. \textit{Analyzer} returns index $ijk$ of the next voxel intersected by the ray($o, d$), and interval of $t \in \lbrack t_0, t_1 \rbrack$ where $o + td$ is inside the voxel. \textit{grid.at} returns whether the voxel at $ijk$ is occupied or not.}
    \begin{algorithmic}[1]
    \Function {dda\_branch}{$ray, t_{\min}, t_{\max}$}
        \State $buf \gets \lbrack\ \rbrack$
        \State $t_{last} \gets t_{\min}$
        \While {$t_{last} \leq t_{\max}$}
            \State // Analyzer is DDA or HDDA.
            \State $ijk, t_0, t_1 \gets \mathrm{Analyzer(}ray, t_{last}\mathrm{)}$
            \While {$t_{last} \leq t_0$}
                \State $t_{last} \gets t_{last} + \Delta t$
            \EndWhile
            \While {$t_{last} \leq t_1$}
            \If {grid.at($ijk$)} // Grid is dense or VDB.
                \State $buf.append(t_{last})$
            \EndIf
                \State $t_{last} \gets t_{last} + \Delta t$
            \EndWhile
        \EndWhile
        \State \Return $buf$
    \EndFunction
    \end{algorithmic}
    \label{alg:branch}
\end{algorithm}

\begin{algorithm}[tb]
    \caption{Algorithm of DDA skip.}
    \begin{algorithmic}[1]
    \Function {dda\_skip}{$ray, t_{\min}, t_{\max}$}
        \State $buf \gets \lbrack\ \rbrack$
        \State $t_{last} \gets t_{\min}$
        \While {$t_{last} \leq t_{\max}$}
            \State $ t_1 \gets t_{\min} $
            \Repeat
                \State $ijk, t_0, t_1 \gets \mathrm{Analyzer(}ray, t_1\mathrm{)}$
            \Until {!grid.at($ijk$)}
            \While {$t_{last} \leq t_0$}
                \State $t_{last} \gets t_{last} + \Delta t$
            \EndWhile
            \While {$t_{last} \leq t_1$}
                \State $buf.append(t_{last})$
                \State $t_{last} \gets t_{last} + \Delta t$
            \EndWhile
        \EndWhile
        \State \Return $buf$
    \EndFunction
    \end{algorithmic}
    \label{alg:skip}
\end{algorithm}

We show the ray-sampling algorithms tested by our work in \cref{alg:branch} and \cref{alg:skip}.
As we describe in \cref{sec:method}, \textit{DDA branch} (\cref{alg:branch}) synchronizes the voxel traversal on all workers, and branches according to each voxel's occupancy. On the other hand, \textit{DDA skip} (\cref{alg:skip}) skips all leading empty voxels and then marches in the occupied voxel. DDA in the both algorithms can be replaced with HDDA.

\end{document}